\newcommand{\CLASSINPUTtoptextmargin}{19.1mm}
\newcommand{\CLASSINPUTbottomtextmargin}{19.1mm}
\newcommand{\CLASSINPUTinnersidemargin}{19.1mm}
\newcommand{\CLASSINPUToutersidemargin}{19.1mm}
\documentclass[letterpaper,10pt,conference,twocolumn,dvipsnames]{IEEEtran}
\usepackage[export]{adjustbox}
\usepackage{algorithm}
\usepackage[%
    indLines=false,
    italicComments=true,
    noEnd=false,
]{algpseudocodex}
\usepackage{amsmath}
\usepackage[font=footnotesize,labelfont=bf]{caption}
\usepackage{circledsteps}
\usepackage{csquotes}
\usepackage{datetime2}
\usepackage{fancyhdr}
\usepackage[%
    toc, % puts the link in the ToC
    record, % to use bib2gls
    abbreviations, % to load abbreviations / acronyms
    nonumberlist, % to avoid printing the numbers of the references in the acronyms page
]{glossaries-extra}
\usepackage{graphicx}
\usepackage{hyperref}
\usepackage{import}
\usepackage{makecell}
\usepackage[final]{microtype}
\usepackage{multirow}
\usepackage{numprint}
\usepackage{pifont}
\usepackage{newtxmath}  % it should substitute amssymb and amsfonts
\usepackage{siunitx}
\usepackage{stfloats}
\usepackage{svg}
\usepackage{tabularx}
\usepackage{textcomp}
\usepackage{tikz}
\usepackage{xcolor}

\usepackage[%
    backend=biber,%
    backref=false,%
    citestyle=numeric,%
    dashed=false,%
    date=iso,%
    doi=false,%
    giveninits=true,%
    hyperref=true,%
    isbn=false,%
    maxbibnames=1,%
    maxcitenames=1,%
    minbibnames=1,%
    mincitenames=1,%
    seconds=true,%
    sortcites=true,%
    style=ieee,%
    sorting=none,%
    url=false,%
    urldate=iso,%
]{biblatex}

\usepackage{enumitem}

\hypersetup{
    colorlinks=true,
    citecolor=.,
    linkcolor=.,
    urlcolor={ForestGreen},
}

\svgsetup{%
    inkscapearea=drawing,
    inkscapepath=svgsubdir,
    inkscapeformat=pdf, % to force usage of PDF
    inkscapelatex=false, % to disable latex rendering of text, produces errors
    inkscapeopt=-T,  % in this way, we pass the option to export the text as path, limiting compatibility issues
}

\svgpath{{images/vector/}}

\graphicspath{{images/raster/}{images/vector/exported-pdf/}}

\npthousandsep{\,}

\newglossary[]{ignored}{}{}{Ignored}
\newglossary[]{conference}{}{}{Conferences}
\newglossary[]{journal}{}{}{Journals}
\newglossary[]{terminology}{}{}{Terminology}

\newabbreviation[
    type={ignored}
]{eo}{EO}{expected outcome}
\newabbreviation[
    type={ignored}
]{nyuad}{NYU Abu Dhabi}{New York University Abu Dhabi, Abu Dhabi, UAE}
\newabbreviation[
    type={ignored}
]{pe}{PE}{proficiency evaluation}
\newabbreviation[
    type={ignored}
]{rq}{RQ}{research question}
\newabbreviation[
    type={ignored}
]{so}{SO}{scientific objective}
\newabbreviation[
    type={ignored}
]{tuwien}{TUWien}{Technische Universit\"at Wien, Vienna, Austria}

\newabbreviation[
    type={terminology}
]{asic}{ASIC}{application specific integrated circuit}
\newabbreviation[
    type={terminology}
]{cnn}{CNN}{convolutional neural network}
\newabbreviation[
    type={terminology}
]{cpu}{CPU}{central processing unit}
\newabbreviation[
    type={terminology}
]{dl}{DL}{deep learning}
\newabbreviation[
    type={terminology}
]{dmr}{DMR}{double modular redundancy}
\newabbreviation[
    type={terminology}
]{dnn}{DNN}{deep neural network}
\newabbreviation[
    type={terminology}
]{fit}{FIT}{failure in time}
\newabbreviation[
    type={terminology}
]{gpu}{GPU}{graphics processing unit}
\newabbreviation[
    type={terminology}
]{iot}{IoT}{internet of things}
\newabbreviation[
    type={terminology}
]{lsb}{LSB}{least significant bit}
\newabbreviation[
    type={terminology}
]{ml}{ML}{machine learning}
\newabbreviation[
    type={terminology}
]{msb}{MSB}{most significant bit}
\newabbreviation[
    type={terminology}
]{nn}{NN}{neural network}
\newabbreviation[
    type={terminology}
]{relu}{ReLU}{rectified linear unit}
\newabbreviation[
    type={terminology}
]{risc}{RISC}{reduced instruction set computer}
\newabbreviation[
    type={terminology}
]{simd}{SIMD}{single instruction multiple data}
\newabbreviation[
    type={terminology}
]{sota}{SotA}{state of the art}
\newabbreviation[
    type={terminology}
]{snn}{SNN}{spiking neural network}
\newabbreviation[
    type={terminology}
]{svm}{SVM}{support vector machine}
\newabbreviation[
    type={terminology}
]{tmr}{TMR}{triple modular redundancy}
\newabbreviation[
    type={terminology}
]{tpu}{TPU}{tensor processing unit}
\newabbreviation[
    type={terminology}
]{vae}{VAE}{variational auto encoder}

\newabbreviation[
    type={conference}
]{dac}{DAC}{Design Automation Conference}
\newabbreviation[
    type={conference}
]{date}{DATE}{Design Automation and Test in Europe Conference and Exhibition}
\newabbreviation[
    type={conference}
]{dsn}{DSN}{International Conference on Dependable Systems and Networks}
\newabbreviation[
    type={conference}
]{iccad}{ICCAD}{International Conference on Computer-Aided Design}
\newabbreviation[
    type={conference}
]{iclr}{ICLR}{International Conference on Learning Representations}
\newabbreviation[
    type={conference}
]{icml}{ICML}{International Conference on Machine Learning}
\newabbreviation[
    type={conference}
]{ijcnn}{IJCNN}{IEEE International Joint Conference on Neural Networks}
\newabbreviation[
    type={conference}
]{iros}{IROS}{IEEE/RJS International Conference on Intelligent Robots and Systems}
\newabbreviation[
    type={conference}
]{nips}{NIPS}{Neural Information Processing Systems}
\newabbreviation[
    type={journal}
]{dandt}{D\&T}{IEEE Design \& Test}
\newabbreviation[
    type={journal}
]{jmlr}{JMLR}{Journal of Machine Learning Research}
\newabbreviation[
    type={journal}
]{tcad}{TCAD}{IEEE Transaction on Computer-Aided Design of Integrated Circuits and Systems}
\newabbreviation[
    type={journal}
]{tecs}{TECS}{ACM Transactions on Embedded Computing Systems}
\newabbreviation[
    type={journal}
]{tvlsi}{TVLSI}{IEEE Transactions on Very Large Scale Integration (VLSI) Systems}

\glsaddall

\glssetcategoryattribute{abbreviation}{glossdesc}{title}

\newcommand{\emphasis}[1]{\emph{#1}}

\newcommand{\workname}{enpheeph}

\newcommand{\emphasizedworkname}{\emphasis{\workname}}

\newcommand{\customcheckmark}{\ding{51}}
\newcommand{\customxmark}{\ding{55}}
\newcommand*\colourmark[3]{%
  \expandafter\newcommand\csname #3\endcsname{\textcolor{#1}{#2}}}
\newcommand{\red}[1]{\textcolor{red}{#1}}

\makeatletter
\newcommand{\printfnsymbol}[1]{%
    \textsuperscript{\@fnsymbol{#1}}%
}
\makeatother

\makeatletter
\newcommand*{\inlineequation}[2][]{%
  \begingroup
    \refstepcounter{equation}%
    \ifx\\#1\\%
    \else
      \label{#1}%
    \fi
    \relpenalty=10000 %
    \binoppenalty=10000 %
    \ensuremath{%
      #2%
    }%
    ~\@eqnnum
  \endgroup
}
\makeatother

\definecolor{blanchedalmond}{rgb}{1.0, 0.92, 0.8}
\definecolor{champagne}{rgb}{0.97, 0.91, 0.81}
 	\definecolor{beige}{rgb}{0.96, 0.96, 0.86}
 	\definecolor{carnelian}{rgb}{0.7, 0.11, 0.11}
\definecolor{crimson}{rgb}{0.86, 0.08, 0.24}
\definecolor{yellowcirclefill}{RGB}{255, 246, 221}
\definecolor{redcircleborder}{RGB}{192, 0, 0}

\newcommand{\BlackCircled}[1]{\Circled[inner color=white,outer color=black,fill color=black]{#1}}
\newcommand{\GoodBlackCircled}[1]{\BlackCircled{\textbf{\small{#1}}}}
\newcommand{\RedCircled}[1]{\Circled[inner color=redcircleborder,outer color=redcircleborder,fill color=yellowcirclefill]{#1}}
\newcommand{\GoodRedCircled}[1]{\RedCircled{\textbf{\small{#1}}}}
\newcommand{\titlecaseabbreviation}[1]{\GlsXtrIfUnusedOrUndefined{#1}{\glsentrytitlecase{#1}{long}\space(\glsxtrshort{#1})\glsunset{#1}}{\glsxtrshort{#1}\glsunset{#1}}}
\newcommand{\titlecaseabbreviationpl}[1]{\GlsXtrIfUnusedOrUndefined{#1}{\glsentrytitlecase{#1}{longpl}\space(\glsxtrshortpl{#1})\glsunset{#1}}{\glsxtrshortpl{#1}\glsunset{#1}}}

\let\svthefootnote\thefootnote
\newcommand\blankfootnote[1]{%
  \let\thefootnote\relax\footnotetext{#1}%
  \let\thefootnote\svthefootnote%
}
\let\svfootnote\footnote
\renewcommand\footnote[2][?]{%
  \if\relax#1\relax%
    \blankfootnote{#2}%
  \else%
    \if?#1\svfootnote{#2}\else\svfootnote[#1]{#2}\fi%
  \fi
}

\newcommand{\setuptoappearheader}[1]{%
    \fancypagestyle{toappearfirstpageheader}{% Page style for first page
        \fancyhf{}% Clear header/footer
        \chead{#1}  % centered header, there is the same for l/r and foot
    }
}

\newcommand{\toappearheader}{%
    \thispagestyle{toappearfirstpageheader}
}

\let\svfootnote\footnote

\let\svfootnote\footnote

\renewcommand{\IEEEtitletopspaceextra}{0pt}

\def\titlefontsize{\fontsize{24.88}{29.856}\selectfont}

\renewcommand{\IEEEtitletopspaceextra}{6mm}

\begin{document}

% to define the checkmark and the xmark
\colourmark{green}{\customcheckmark}{greencheckmark}
\colourmark{red}{\customxmark}{redxmark}

% by default it should be \Huge
\title{\titlefontsize enpheeph: A Fault Injection Framework for Spiking and Compressed \glsentrytitlecase{dnn}{longpl}}

% \thanks does not work here, use \blankfootnote after for referring to numbers/symbols
\author{%
\IEEEauthorblockN{Alessio Colucci$^{1}$, Andreas Steininger$^{1}$, Muhammad Shafique$^{2}$}
\IEEEauthorblockA{\textit{$^1$Institute of Computer Engineering, Technische Universit{\"a}t Wien, Vienna, Austria}} 
\IEEEauthorblockA{\textit{$^2$eBrain Lab, Division of Engineering, New York University Abu Dhabi, UAE}}\vspace*{-4mm}\\
Email: \{alessio.colucci,andreas.steininger\}@tuwien.ac.at,muhammad.shafique@nyu.edu\\
}

% \thanks does not work here, use \blankfootnote after for referring to numbers/symbols
% \author{\IEEEauthorblockN{Alessio Colucci$^{1}$\printfnsymbol{1}\thanks{\printfnsymbol{1}These authors contributed equally}, Alberto Marchisio$^{1}$, Beatrice Bussolino$^{2}$, Vojtech Mrazek$^3$,\\Maurizio Martina$^2$, Guido Masera$^2$, Muhammad Shafique$^{1,4}$}
% \IEEEauthorblockA{\textit{$^1$Technische Universit{\"a}t Wien, Vienna, Austria}} 
% \IEEEauthorblockA{\textit{$^2$Politecnico di Torino, Turin, Italy}}
% \IEEEauthorblockA{\textit{$^3$Faculty of Information Technology, IT4Innovations Centre of Excellence, Brno University of Technology, Czech Republic}}
% \IEEEauthorblockA{\textit{$^4$Division of Engineering, New York University Abu Dhabi, UAE}}\vspace*{-3mm}\\
% Email: \{alessio.colucci,alberto.marchisio,muhammad.shafique\}@tuwien.ac.at,muhammad.shafique@nyu.edu\\
% \{beatrice.bussolino,maurizio.martina,guido.masera\}@polito.it,mrazek@fit.vutbr.cz\vspace*{-3mm}\\
% }

\maketitle
% to setup the header with the line and the to appear phrase
\setuptoappearheader{To appear at 2022 IEEE/RSJ International Conference on Intelligent Robots and Systems (IROS), October, 2022}
\toappearheader

\begin{abstract}
\glslocalresetall Research on \titlecaseabbreviationpl{dnn} has focused on improving performance and accuracy for real-world deployments, leading to new models, such as \titlecaseabbreviationpl{snn}, and optimization techniques, e.g., quantization and pruning for compressed networks. 
However, the deployment of these innovative models and optimization techniques introduces possible reliability issues, which is a pillar for \glspl{dnn} to be widely used in safety-critical applications, e.g., autonomous driving. Moreover, scaling technology nodes have the associated risk of multiple faults happening at the same time, a possibility not addressed in state-of-the-art resiliency analyses.

% Current state-of-the-art frameworks for resiliency analysis are tailored to analyze distinctive fault patterns happening in specific \glspl{dnn}, not allowing the analysis of newer models and techniques without long implementation delays. In addition, they lack optimized operations and support for non-standard fault patterns, which are required to make the analyses future-proof.

Towards better reliability analysis for \glspl{dnn}, we present
\emphasizedworkname, a Fault Injection Framework for Spiking and Compressed \glspl{dnn}. The \emphasizedworkname~framework enables optimized execution on specialized hardware devices, e.g., \glsxtrshortpl{gpu}, while providing complete customizability to investigate different fault models, emulating various reliability constraints and use-cases. Hence, the faults can be executed on \glspl{snn} as well as compressed networks with minimal-to-none modifications to the underlying code, a feat that is not achievable by other state-of-the-art tools.

To evaluate our~\emphasizedworkname~framework, we analyze the resiliency of different \gls{dnn} and \gls{snn} models, with different compression techniques. By injecting a random and increasing number of faults, we show that \glspl{dnn} can show a reduction in accuracy with a fault rate as low as $7 \times 10 ^{-7}$ faults per parameter, with an accuracy drop higher than $40\%$.
% Additionally, we show the increased resiliency of \glspl{snn}, which do not show any accuracy reduction with fault rates as high as $6 \times 10 ^{-2}$.
Run-time overhead when executing \emphasizedworkname~is less than $20\%$ of the baseline execution time when executing $\numprint{100000}$ faults concurrently, at least $10 \times$ lower than state-of-the-art frameworks, making \emphasizedworkname~future-proof for complex fault injection scenarios.

We release the source code of our \emphasizedworkname~framework under an open-source license at \url{https://github.com/Alexei95/enpheeph}.

\glslocalresetall

\end{abstract}

% Note that keywords are not normally used for peer-reviewed papers.

\glsresetall

% \import{content/paper/sections/}{0_round0}

\section{Introduction}
\label{section:introduction}

In the last decade, \titlecaseabbreviationpl{dnn} have seen an exponential increase in practical applications~\cite{abiodunStateoftheartArtificialNeural2018,dongSurveyDeepLearning2021}, due to their ability to learn complex patterns beyond classical hard-coded algorithms. A possible application is autonomous driving, which is becoming more prominent at different capability levels~\cite{on-roadautomateddrivingoradcommitteeTaxonomyDefinitionsTerms}. However, strong error-tolerance and resiliency are required to reach high autonomous capability levels, as detailed in ISO 26262~\cite{internationalorganizationforstandardizationISO26262102018}, which indicates a \titlecaseabbreviation{fit} rate of fewer than 100 failures in 1 billion hours of operation for the highest safety level. The exponential increase of multiple upset events in advanced technology nodes~\cite{blackPhysicsMultipleNodeCharge2013,nealeNeutronRadiationInduced2016}, makes this threshold complex to achieve and maintain.

Even though resiliency to faults and errors is of foremost importance, there have been few in-depth resiliency analyses for the effect of faults on \glspl{dnn}. Some examples focus on permanent faults~\cite{zhangThundervoltEnablingAggressive2018,ozenBoostingBitErrorResilience2020,reagenAresFrameworkQuantifying2018,hoangTReMapReducingOverheads2021,hoangFTClipActResilienceAnalysis2020,hanifDNNLifeEnergyEfficientAging2021}, while others only consider \titlecaseabbreviationpl{cnn}~\cite{liUnderstandingErrorPropagation2017,liTensorFIConfigurableFault2018,chenBinFIEfficientFault2019,chenLowcostFaultCorrector2021}. These tools focus on analyzing a single fault happening at a certain time inside the model, a fault model which is bound to be superseded by multiple fault events due to the aforementioned technology scaling. Hence, state-of-the-art tools are not optimized for scalability, making it very difficult to inject multiple faults in the models without affecting the run-time.

Additionally, many new techniques and architectures for improving \gls{dnn} efficiency have been developed, such as quantization~\cite{hanLearningBothWeights2015}, pruning~\cite{hanLearningBothWeights2015} or \titlecaseabbreviationpl{snn}~\cite{maassNoisySpikingNeurons1996,maassNetworksSpikingNeurons1997,putraSoftSNNLowCostFault2022,wicaksanaputraReSpawnEnergyEfficientFaultTolerance2021}, making them more challenging to analyze using traditional methodologies. As state-of-the-art tools are tailored to specific platform/model configurations, it proves that it is difficult to quickly adapt them to the constantly-evolving model space and optimization techniques.

\subsection{Motivational Case Study}

\begin{figure}[t]
    \centering
    \includegraphics[width=\linewidth]{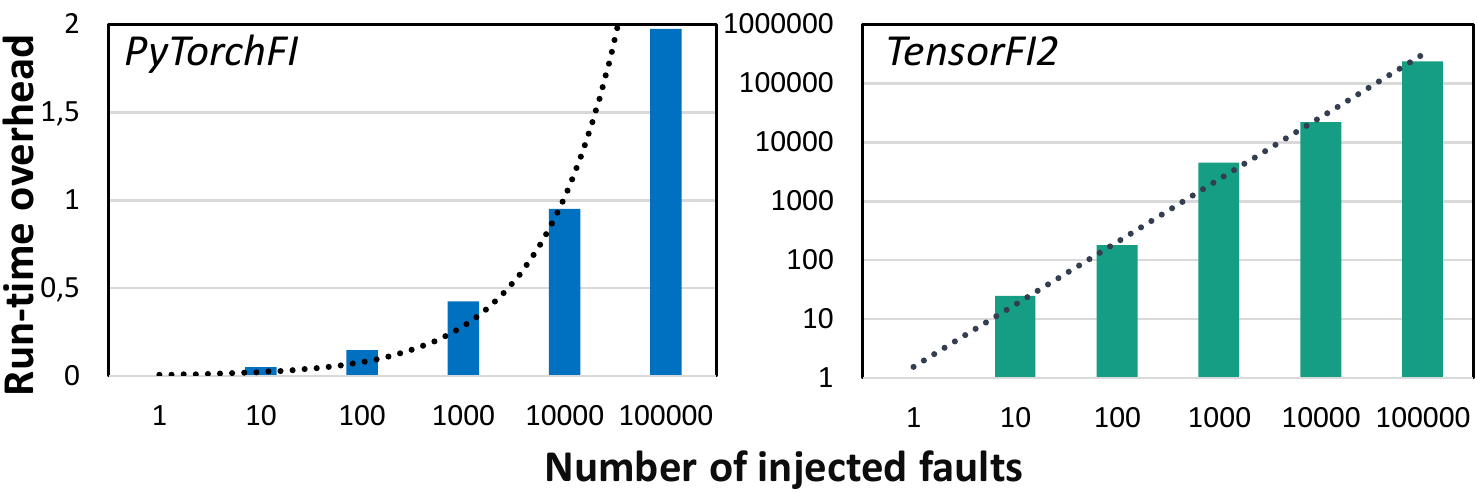}
    \caption{Comparative analysis of run-time overhead between different tools. The overhead is measured in multiplicative units compared to the baseline, so an overhead of 0.5 indicates an execution time that is 1.5$\times$ the fault-free baseline execution time. PyTorchFI, on the left, has the ordinate axis in linear coordinates, while TensorFI2 on the right uses a logarithmic ordinate axis, hence the linear trendline with an exponential pattern.} We can note the exponential growth pattern against the number of concurrently injected faults.
    \label{figure:motivation}
\end{figure}

We show a comparison of run-time overhead when using different state-of-the-art fault injection frameworks in Fig.~\ref{figure:motivation}. By running from \numprint{1} to \numprint{100000} injections, we can see how much overhead is incurred when using multiple frameworks, which grows exponentially. Hence, using these frameworks for multiple faults makes injection experiments very slow, which affects the system design phase.
Additionally, these frameworks are not easily adaptable to new technologies or different deep learning libraries, as their code is tied to the specific framework and neural network on which they are implemented.
Hence, we can see how a scalable and adaptable framework is necessary for making resiliency analysis of \glsxtrshortpl{dnn} future-proof.

\subsection{Research Questions}

The aforementioned case study leads us to formulate the following research questions:

\begin{itemize}
    \item How can we maintain performance when executing multiple fault injections simultaneously?
    \item How can we develop a generic fault injection framework capable of adapting to different models with minimal modifications?
    \item How can we carry out the resiliency analysis for \glspl{snn} and compressed \glspl{dnn}?
\end{itemize}

\subsection{Novel Contributions}

To answer the research questions, we provide the following novel contributions:

\begin{itemize}
    \item we develop \emphasizedworkname{}, a modern fault injection framework, capable of handling multiple fault injections with minimum overhead, and adaptable to all models and configurations with minimal-to-none modifications;
    \item we release \emphasizedworkname{} under an open-source license at \url{https://github.com/Alexei95/enpheeph};
    \item we employ \emphasizedworkname{} to analyze the resiliency of different \glspl{dnn}, as well as \gls{snn} for gesture recognition, employing different compression techniques;
\end{itemize}

After a brief background and related work analysis, in Sections~\ref{section:background} and~\ref{section:related_work}, we discuss the methodology and the implementation behind our \emphasizedworkname{} framework in Section~\ref{section:methodology}. Then, we show our experimental setup in Section~\ref{section:experimental_setup}, and analyze the fault injection results in Section~\ref{section:evaluation}. We draw the conclusion on our work in Section~\ref{section:conclusion}.

\section{Background}
\label{section:background}

\subsection{Fault Injection for Neural Networks}

Fault injection is used to test the behaviour of a system when an unexpected state is erroneously reached.
Faults are classified mainly into two types, transient, which disappear after a concise time interval, and permanent, which are not repairable. Also, depending on the outcome of the affected signals, they are categorized as bit-flip if the signal value is inverted or stuck-at if the signal value is stuck at a 0 or 1 in bit value.
Our focus will be on transient faults, which are caused by particles interacting with the hardware and flipping the signal values.
In the case of neural networks, these can happen in different locations. However, when using software-level injection methodologies, only a limited set of faults can be covered without prior knowledge of the inner workings of the software and hardware platforms. This means the only directly addressable elements are the tensor values and their indices for layer weights and outputs, representing faults happening in memory locations for the weights and the temporary layer outputs.

\subsection{Spiking Neural Networks}

\begin{figure}[H]
    \centering
    \includegraphics[width=\linewidth]{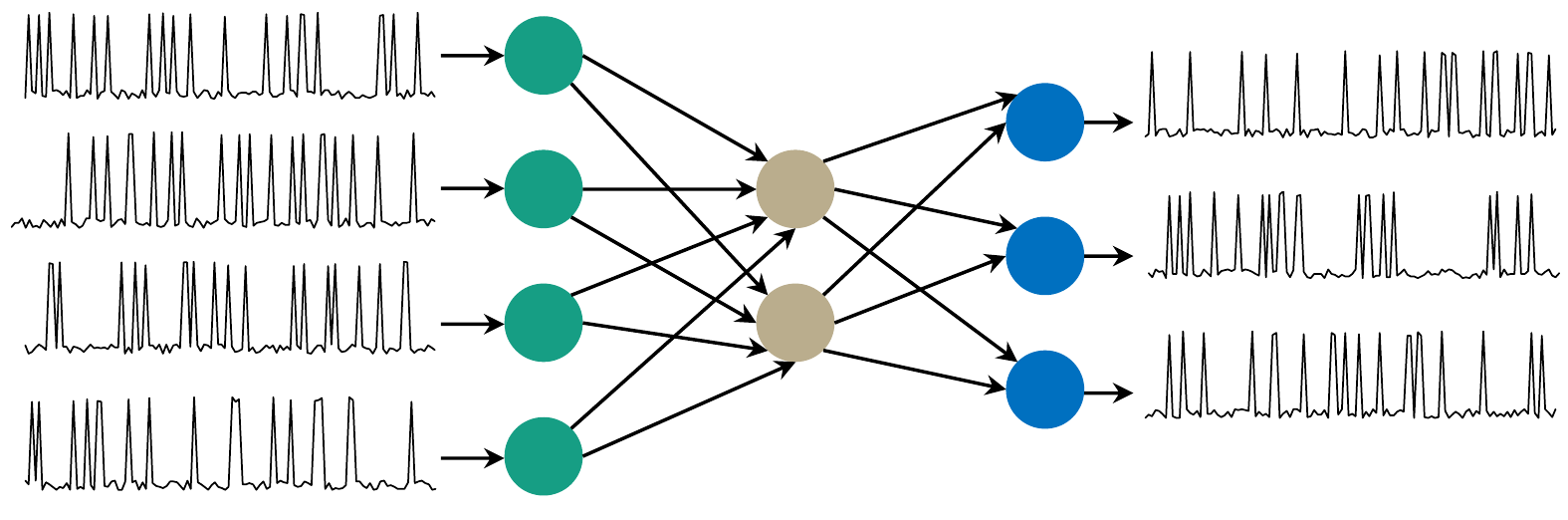}
    \caption{\glsentrytitlecase{snn}{longpl} can accommodate the same model structures of \glsxtrshortpl{dnn}, but each neuron processes a stream of events evolving over time.}
    \label{figure:snn_background}
\end{figure}

Along the line of brain-inspired neural networks, \glslocalreset{snn}\titlecaseabbreviationpl{snn} have been developed recently~\cite{maassNoisySpikingNeurons1996,maassNetworksSpikingNeurons1997}, which use temporal correlation and increase the amount of information learned from the inputs, due to the differential equation that governs the neuron outputs.
In this way, fewer data inputs are required to reach a similar accuracy as of standard \glsxtrshortpl{dnn}~\cite{sorbaroOptimizingEnergyConsumption2020}, while the complexity of each input is higher.
This increase in input complexity leads to increased computational complexity, which also opens possibilities for resiliency issues.
The main difference between a standard artificial neuron and a spiking neuron is that in the latter, the output is driven by a differential equation, which encodes additional information based on the time correlation of the inputs, as can be seen in Figure~\ref{figure:snn_background}.
\glspl{snn} have seen an enormous increase in practical applications, especially for complex tasks~\cite{schliebsEvolvingSpikingNeural2013,bingSurveyRoboticsControl2018}. However, they are still not well-integrated in the deep learning frameworks; hence, their development is based on custom hardware accelerators~\cite{bouvierSpikingNeuralNetworks2019}.

% \begin{enumerate}
%     \item Along the line of brain neuronal networks, \glslocalreset{snn}\titlecaseabbreviationpl{snn} have been developed recently~\cite{maassNoisySpikingNeurons1996,maassNetworksSpikingNeurons1997}, to use temporal correlation and increase the information amount learned from the inputs.
%     \item In this way, fewer data inputs are required to reach a similar amount of accuracy~\cite{sorbaroOptimizingEnergyConsumption2020}, while the complexity of each input is higher.
    
%     \item Additionally, this increase in data density is compensated with a huge increase in computational complexity, which also opens possibilities for resiliency issues.
    
%     \item The main difference between a classical artificial neuron and a spiking neuron is that the output is driven by a differential equation, which encodes additional information based on the time correlation of the inputs.
    
%     \item \glspl{snn} have seen an enormous increase in practical applications, especially for complex tasks \red{REFS}. However, they are still not well-integrated in the deep learning frameworks, hence their development has mostly been based around custom hardware accelerators.
% \end{enumerate}

\subsection{Compression Techniques}

Many software techniques are developed to optimize the \glsxtrshort{dnn} requirements while maintaining similar accuracy, leading to compressed networks, where their memory and computational requirements are lower than their original counterparts.

\begin{figure}[H]
    \centering
    \includegraphics[width=\linewidth]{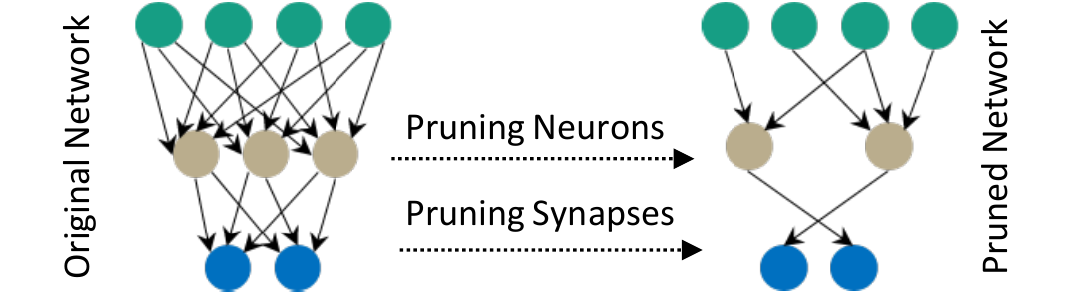}
    \caption{Pruning can be used to remove redundant neurons and synapses, providing similar accuracy with smaller networks.}
    \label{figure:pruning_background}
\end{figure}

An example of these techniques is pruning~\cite{hanLearningBothWeights2015}, which reduces the number of neuron connections, i.e., synapses or neurons, as shown in Fig.~\ref{figure:pruning_background}. However, it is most effective when the underlying hardware supports sparse execution~\cite{qinSIGMASparseIrregular2020}, such that the operations can be executed on smaller matrices containing the coordinates and the values of the elements, therefore representing only the non-zero elements~\cite{CoordinateFormatCOO}.

On the other hand, quantization~\cite{gholamiSurveyQuantizationMethods2021} reduces the number of bits used to represent the data in the inputs, the outputs, and the weights. It is used during deployment to reduce 32-bit floating-point numbers to  8-bit integers with negligible loss in accuracy. In addition to reducing the memory footprint, the operations can be optimized further, allowing parallel operations to run faster than 32-bit floating-point numbers.

\section{Related Work}
\label{section:related_work}

We compare existing state-of-the-art frameworks in Table~\ref{table:sota_injection_framework}, also covering the features provided by \emphasizedworkname{}.
Compared to the other state-of-the-art frameworks, \emphasizedworkname{} is not a collection of scripts, but it has been developed as a homogeneous set of software API primitives. With this approach, \emphasizedworkname{} can be easily adapted to different underlying libraries providing different functionalities, without having to modify its internal code-base. On the other hand, state-of-the-art frameworks do not provide a generic API for easy programmability, but simply a set of configuration knobs which can be accessed via predefined configuration files. This configuration method is limiting as it does not allow for configurations beyond the original scopes of the work.
The easy-to-use API additionally enables any new user of the \emphasizedworkname{} framework to write their own customized injection while leveraging the setup/restore backbones provided with \emphasizedworkname{}. An example of this easy implementation is related to different hardware devices, as the original implementation covered only CPU execution, but was programmatically extended to GPU compatibility with almost zero code changes, while state-of-the-art frameworks require extensive updates to allow execution on GPU.

Of all the frameworks, only \emphasizedworkname{} has no implementation details tied with the underlying \glsxtrshort{dnn} framework, making it adaptable to different \glsxtrshort{dnn} frameworks with minimal-to-none internal code changes.
Moving on to the targetability of the elements, \emphasizedworkname{} allows for custom targetability from the bit-level through the tensor-level to the layer-level, allowing a custom number of bit-precise injections during an execution. The state-of-the-art frameworks are limited to a single layer, and there are limits on executing a single specific fault or a random one over the whole tensor.
\emphasizedworkname{} provides customizability for the fault and the target type, while providing implementations for the basic fault types, e.g., bit-flips and stuck-at. Other frameworks support only bit-flips, with the only exception being TensorFI2, also supporting stuck-at faults, and PyTorchFI providing only stuck-at injections.
Similarly, state-of-the-art frameworks allow for targeting layer weights or outputs, but implementing other injections, e.g., on the temporary buffers, is not allowed, while \emphasizedworkname{} easily allows for such a possibility if required.
Moving to compressed networks, \emphasizedworkname{} is the only framework providing full support for sparse tensors and quantization, beyond what is currently offered by \glsxtrshort{dnn} framework, as most of them lack direct support.
Finally, most of the frameworks run only on \glsxtrshort{cpu}, and they are not easily extendable on \glsxtrshort{gpu}, with \emphasizedworkname{} having support for additional devices as well.

Overall, state-of-the-art frameworks are not customizable enough for the evolving scenario of models and techniques, and they do not guarantee multiple fault injection capabilities. \emphasizedworkname{} aims at addressing all of the aforementioned issues.

\begin{table}[h]
    \centering
    % use setlength to change the horizontal spacing between cells
    % it is local, i.e. it must be repeated in each environment
    \setlength{\tabcolsep}{2pt}
    \resizebox{\linewidth}{!}{%
        \begin{tabular}{c|c|c|c|c|c|}
\cline{2-6}
                                                                     & \red{\emphasis{\textbf{enpheeph}}}                               & \textbf{TensorFI2}                     & \textbf{InjectTF2}                     & \textbf{PyTorchFI}                     & \textbf{TorchFI} \\
                                                                     & & \cite{liTensorFIConfigurableFault2018} & \cite{beyerFaultInjectorsTensorFlow2020} & \cite{mahmoudPyTorchFIRuntimePerturbation2020} & \cite{goldsteinReliabilityEvaluationCompressed2020} \\ \hline
\multicolumn{1}{|c|}{\textbf{Library}}                                                         & \begin{tabular}[c]{@{}c@{}}PyTorch\\ Custom\end{tabular}           & TensorFlow2                                             & TensorFlow2                                             & PyTorch                                                 & PyTorch                                                 \\ \hline
\multicolumn{1}{|c|}{\textbf{\begin{tabular}[c]{@{}c@{}}Bit\\ Targetability\end{tabular}}}     & \begin{tabular}[c]{@{}c@{}}Single\\ Multiple\\ Custom\end{tabular} & \begin{tabular}[c]{@{}c@{}}Single\\ Random\end{tabular} & \begin{tabular}[c]{@{}c@{}}Single\\ Random\end{tabular} & Single                                                 & Random                                                  \\ \hline
\multicolumn{1}{|c|}{\textbf{\begin{tabular}[c]{@{}c@{}}Tensor\\ Targetability\end{tabular}}}  & \begin{tabular}[c]{@{}c@{}}Single\\ Multiple\\ Custom\end{tabular} & Random                                                  & Random                                                  & Single                                                  & Random                                                  \\ \hline
\multicolumn{1}{|c|}{\textbf{\begin{tabular}[c]{@{}c@{}}Layer\\ Targetability\end{tabular}}}   & \begin{tabular}[c]{@{}c@{}}Single\\ Multiple\\ Custom\end{tabular} & Single                                                  & Single                                                  & Single                                                  & Single                                                  \\ \hline
% \multicolumn{1}{|c|}{\textbf{\begin{tabular}[c]{@{}c@{}}Supported\\ Operations\end{tabular}}}  & Any                                                                & Any                                                     & Any                                                     & Any                                                     & Limited                                                 \\ \hline
\multicolumn{1}{|c|}{\textbf{\begin{tabular}[c]{@{}c@{}}Fault\\ Type\end{tabular}}}            & \begin{tabular}[c]{@{}c@{}}Bit-flip\\ Stuck-at\\ Custom\end{tabular}   & \begin{tabular}[c]{@{}c@{}}Bit-flip\\ Stuck-at\end{tabular} & Bit-flip                                                    & Stuck-at                                                   & Bit-flip                                                    \\ \hline
\multicolumn{1}{|c|}{\textbf{\begin{tabular}[c]{@{}c@{}}Target\\ Type\end{tabular}}}           & \begin{tabular}[c]{@{}c@{}}Weight\\ Output\\ Custom\end{tabular}   & \begin{tabular}[c]{@{}c@{}}Weight\\ Output\end{tabular} & Output                                                  & \begin{tabular}[c]{@{}c@{}}Weight\\ Output\end{tabular} & \begin{tabular}[c]{@{}c@{}}Weight\\ Output\end{tabular} \\ \hline
\multicolumn{1}{|c|}{\textbf{\begin{tabular}[c]{@{}c@{}}Quantization\\ Support\end{tabular}}}  & Full                                                               & No                                                      & No                                                      & Limited                                                      & Limited                                                 \\ \hline
\multicolumn{1}{|c|}{\textbf{\begin{tabular}[c]{@{}c@{}}Sparse Tensor\\ Support\end{tabular}}} & Full                                                               & No                                                      & No                                                      & No                                                      & Limited                                                 \\ \hline
\multicolumn{1}{|c|}{\textbf{\begin{tabular}[c]{@{}c@{}}Hardware\\ Support\end{tabular}}}      & \begin{tabular}[c]{@{}c@{}}CPU\\ GPU\\ Custom\end{tabular}         & CPU                                                     & CPU                                                     & CPU                                                     & \begin{tabular}[c]{@{}c@{}}CPU\\ GPU\end{tabular}       \\ \hline
\end{tabular}

    }
    \caption{Comparison among the different \glsxtrshort{sota} fault injection frameworks for \glsentrytitlecase{dnn}{longpl}}
    \label{table:sota_injection_framework}
\end{table}

\section{Methodology}
\label{section:methodology}

\begin{figure*}[h!]
    \centering
    \includegraphics[width=\linewidth]{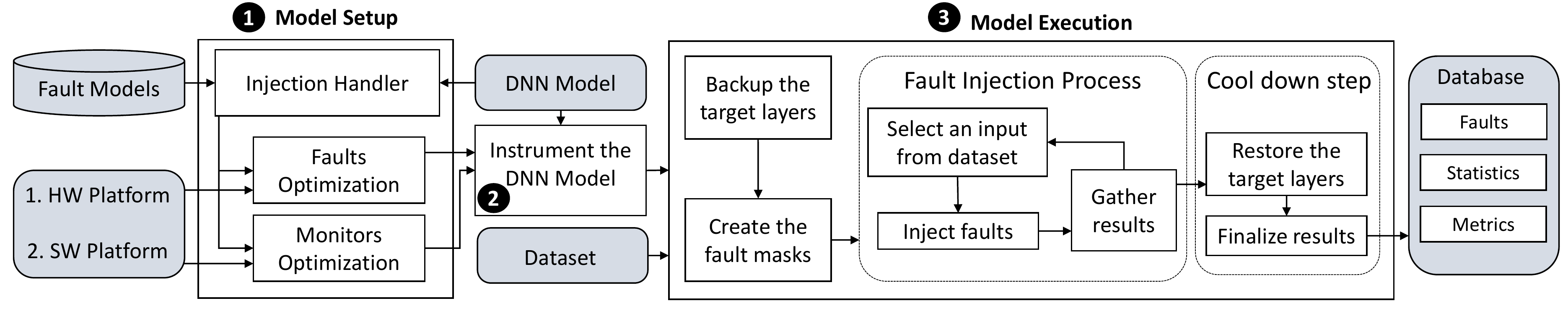}
    \caption{Methodology for our \emphasis{enpheeph} framework. The filled boxes represent the different inputs on the left and the output on the right.}
    \label{figure:methodology}
\end{figure*}

We approach the problem of lightweight fault injection and customizability outlined in Section~\ref{section:introduction}, by developing a generic fault injection framework for \glsxtrshortpl{dnn}, \emphasis{enpheeph}. The methodology is shown in Fig.~\ref{figure:methodology}, and it will be described in detail in the following sub-sections.

\subsection{Inputs}

The \emphasis{enpheeph} framework requires 4 inputs:

\begin{itemize}
    \item \textbf{One or more fault models:} they are required to generate the targets for the faults and the monitors to be inserted in the model.
    \item \textbf{The set of hardware and software platforms:} they are required to employ the correct injection handler and optimized operations, as each handler has a different internal implementation depending on the used \glsxtrshort{dnn} framework.
    \item \textbf{The target \glsxtrshort{dnn} model:} it is used to compute the dimensions of the faults to be injected.
    \item \textbf{The target dataset:} it is required to run the model executions and gather the injection results.
\end{itemize}

\subsection{Model Setup}

In Fig.~\ref{figure:methodology}~\GoodBlackCircled{1} we can see the \enquote{Model Setup} step, which is necessary for setting up the faults and the monitors based on the fault model and the running hardware/software platform.

Each fault represents the set of bits which need to be modified by the mask. Each monitor contains which values must be saved for each execution, as the modified bits are not automatically logged. In this way, there is full customizability of the values which need to be saved.

The \emphasis{enpheeph} framework initially creates an injection handler, which is entitled to generate the faults and the monitors based on the fault model and the dimensions of the target \glsxtrshort{dnn} model. The faults and the monitors are optimized based on the combination of software and hardware platforms, guaranteeing the lowest possible overhead. The instrumented \glsxtrshort{dnn} model, showed in~\GoodBlackCircled{2}, is used for the \enquote{Model Execution} step in the following sub-section.

\subsection{Model Execution}

Continuing with Fig.~\ref{figure:methodology}~\GoodBlackCircled{3}, the instrumented DNN model is executed with the given dataset, meant as a collection of data on which to test, e.g., a subset of the CIFAR10 dataset. All the targeted layers are backed up at the beginning of the execution, and the fault masks are created and placed in the layer execution flows. Then, as shown in the \enquote{Fault Injection Process}, the model is executed with the faults, and the results are gathered based on the monitor configurations. This process is repeated for each input in the dataset. When the dataset is extinguished, the injected layers are restored to their backed-up version, and the results are finalized in the output database.

\subsection{Output}

There is only one output from the \emphasis{enpheeph} framework, and it is a database containing the results of all the monitors, as well as the metrics used for evaluating the model, e.g., accuracy for a classification task.

\subsection{Example Implementation: PyTorch Injection Handler Setup}

\begin{figure}[h]
    \begin{algorithm}[H]
        \begin{footnotesize}
        \captionsetup{font=footnotesize}
        \caption{PyTorch Implementation of the Injection Handler Setup}
        \label{algorithm:methodology_example}
        \begin{algorithmic}[1]
            \LComment{each $injection$ in the input list is either a fault or a monitor}
            \LComment{each $injection$ is a structure containing the necessary information}
            \Procedure{InjectionHandlerSetup}{$model$, $list_{injections}$}
                \For{$injection$ in $list_{injections}$}
                    \LComment{the layer with the same name as in the $injection$ is selected}
                    \State $module$ $\gets$ layer from $model$ using $injection.layerName$
                    
                    \LComment{the correct $target$ is selected based on the injection type}
                    \If{$injection.target$ $==$ $output$}
                        \State $target$ $\gets$ $module.output$
                    \ElsIf{$injection.target$ $==$ $weight$}
                        \State $target$ $\gets$ $module.weight$
                    \EndIf
                    
                    \If{$injection.type$ $==$ $fault$}
                        \LComment{if we have a fault we create a $mask$}
                        \LComment{from the type of fault and}
                        \LComment{expand it to the $target$ size}
                        \State $maskElement$ $\gets$ $injection.faultType$ \label{algorithm:line:mask_creation}
                        \State $mask$ $\gets$ expand $maskElement$ to $target.shape$
                        \LComment{an execution hook is added to the $module$}
                        \LComment{to update the $target$}
                        \LComment{the update involves the $mask$ to force the injection}
                        \State add exec. hook in $module$, $target$ $\gets$ $target + mask$ \label{algorithm:line:hook_fault}
                    \ElsIf{$injection.type$ $==$ $monitor$}
                        \LComment{a similar hook is added to the $module$}
                        \LComment{if we are running a monitor,}
                        \LComment{in order to save the $target$}
                        \State add execution hook in $module$, to save $target$ \label{algorithm:line:hook_monitor}
                    \EndIf
                \EndFor
            \EndProcedure
        \end{algorithmic}
        \end{footnotesize}
    \end{algorithm}
\end{figure}

In Algorithm~\ref{algorithm:methodology_example}, we show an example for the methodology implementation in the \emphasizedworkname{} framework: the PyTorch \cite{paszkePyTorchImperativeStyle2019} implementation of the setup phase of the injection handler.

This procedure requires the list of injections, where each injection is a structure containing the information of the target layer name, the target type, the fault type and the indices of the target elements.

After selecting the correct layer from the layer name in the injection, the correct target is selected, choosing between output and weight. Then, depending on whether the injection is a fault or a monitor, the mask creation process at line \ref{algorithm:line:mask_creation} consists of expanding the bit-wise mask to the shape of the target array, so it can be combined with the target at runtime. Finally, we add an execution hook to either update the target with the mask at line \ref{algorithm:line:hook_fault} or save the target at line \ref{algorithm:line:hook_monitor}, if the injection is a fault or a monitor respectively.

If we are injecting a fault, we have developed an automated interface capable of determining on which device type we are operating, e.g. CPU vs GPU vs TPU. In this way the mask injection is tailored to the specific device, employing different back-end libraries and further accelerating the execution.

In this particular implementation, we do not require backing up the layer before the injection handler setup. PyTorch allows using execution hooks that do not leave permanent modifications on the model once removed. The execution hooks are run right before or after the execution of the layer, depending on the chosen hook, allowing for updates at runtime before the execution starts. This enables further speed-ups and memory savings when compared to the complete backup and restore processes of the target layers.

\section{Experimental Setup}
\label{section:experimental_setup}

To evaluate our \emphasis{enpheeph} framework, we implement the injection handler and the corresponding components detailed in Section~\ref{section:methodology}. We also train and test multiple models and fault rates on different hardware platforms. All the trained models and their configurations will be available in the open-source release of the framework.

\begin{figure}[h]
    \centering
    \includegraphics[width=\linewidth]{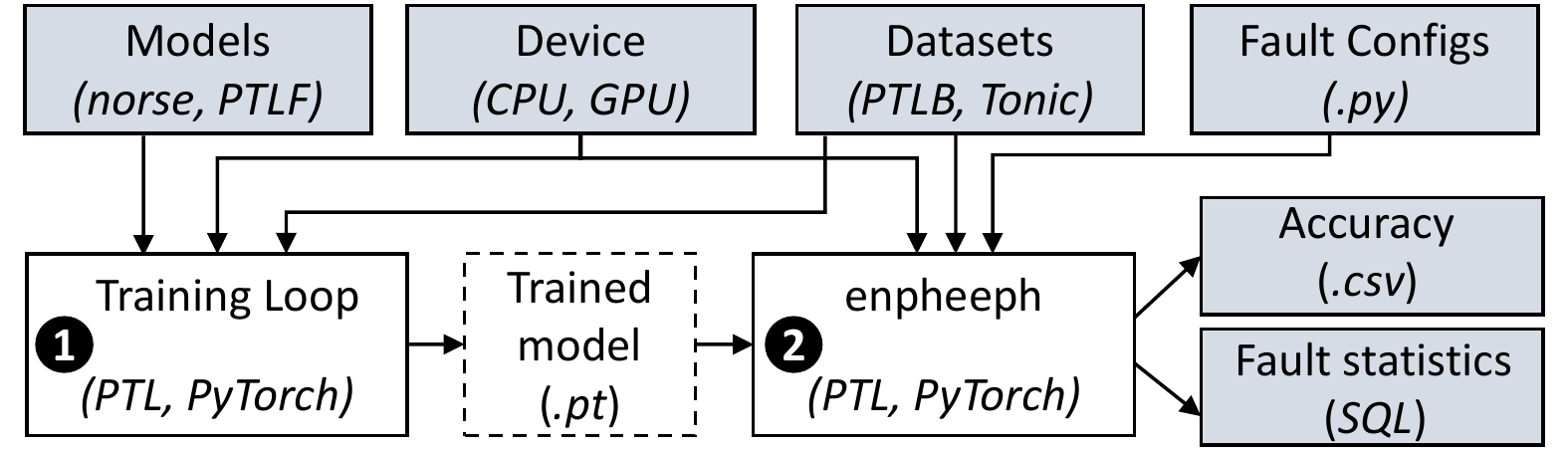}
    \caption{Overview of the used experimental setup for training the \glsxtrshort{dnn} models and for testing our \emphasizedworkname{} framework.}
    \label{figure:experimental_setup}
\end{figure}

An overview of the experimental setup is given in Fig.~\ref{figure:experimental_setup}, where we can see the training loop in \GoodBlackCircled{1} depending on the models, the chosen device and the different datasets to produce a trained model. The trained model is then fed as input to a similar testing loop in \GoodBlackCircled{2}, comprising \emphasizedworkname{} and requiring the datasets as well as the fault configurations, to execute the fault injections. Th results are gathered into different file formats, namely a CSV file for the accuracy and all the model metrics, and a SQL database for the results gathered by the monitors and a trace of the faults which have been executed.

\subsection{Hardware}

We test our injections on both CPU and GPU, running on AMD Ryzen Threadripper 2990WX \cite{AMDRyzenThreadripper} and NVIDIA RTX 2080Ti \cite{GraphicsReinventedNVIDIA}.

\subsection{Software}

% \red{I MIGHT ADD AN ALGORITHM AND/OR IMAGE EXPLAINING THE SETUP IN MORE DETAIL}

We implement the low-level compatibility layer of the \emphasis{enpheeph} framework in PyTorch~\cite{paszkePyTorchImperativeStyle2019}. However, the components can be easily implemented on other \glsxtrshort{dnn} libraries.
The injections on the CPU are implemented via NumPy arrays~\cite{harrisArrayProgrammingNumPy2020}, while for the GPU, we use CuPy~\cite{cupy_learningsys2017}.
We additionally use PyTorch Lightning (PTL)~\cite{Falcon_PyTorch_Lightning_2019} to expedite the models' training and testing due to its seamless switch between CPU and GPU execution.
We employ PyTorch Lightning Flash (PTLF)~\cite{PyTorchLightningLightningflash2022} and PyTorch Lightning Bolts (PTLB)~\cite{falcon2020framework} for expediting access to task pipelines, models, and datasets.
For \glsxtrshortpl{snn}, we employ the norse framework~\cite{norse2021} and the tonic dataset collection~\cite{lenz_gregor_2021_5079802} on top of PyTorch for easier \glsxtrshort{snn} implementation.

\subsection{Models}

We choose the VGG11~\cite{simonyanVeryDeepConvolutional2015} and ResNet18~\cite{heDeepResidualLearning2016} models as the backbone for the image classification task on CIFAR10~\cite{Krizhevsky2009LearningML}. Each model is trained with the Adam~\cite{kingmaAdamMethodStochastic2015} optimizer, using 0.001 as the learning rate for up to 60 epochs; the early stopping technique limits the training if the loss starts increasing from the minimum reached value~\cite{girosiRegularizationTheoryNeural1995}.

For the \glsxtrshort{snn}, we develop a shallow network with 2 convolutional layers and a fully-connected layer to be run on the DVS Gesture Dataset~\cite{amirLowPowerFully2017}. It is trained for 25 epochs with the Adam optimizer and 0.001 learning rate.

\subsection{Experiments}
\label{subsection:experiments}

We inject a variable number of faults, determined by the fault rate per number of parameters via sampling of a uniform random distribution.
We choose to sweep the range from $1 \times 10^{-7}$ until $1$, which means we start injecting one fault per ten million parameters until we reach one fault per parameter. For each decade we cover nine different points, e.g. we sweep from $1 \times 10 ^{-7}$ to $9 \times 10 ^{-7}$ via $1 \times 10 ^{-7}$ increases.
For each fault rate we iterate over the model layers and inject the faults one layer at a time, selecting the lowest testing accuracy per each fault rate.

For the compressed networks, there is limited support at the time of writing in PyTorch; hence we target only a limited subset of operations. We inject faults only on the indices of the sparse outputs of each layer. To obtain a sparse representation, we convert the dense output to a sparse index-value representation~\cite{CoordinateFormatCOO}, inject the faults on the index array and convert back to a dense format.
Similarly, to inject faults on quantized networks, we convert the output of each layer to 32-bit integer from 32-bit floating-point, inject the faults, and convert them back to 32-bit floating-point.
To increase the dynamic range in integer representation, as the conversion truncates the original floating-point numbers, we multiply the floating-point by $2^{24}$, so that the original range in floating point is $\left[-128, 127\right]$, while keeping a precision of $\approx 6 \times 10^{-8}$ in integer representation. These numbers were verified to be well-above the range of the tensors in the employed models, hence not affecting the output dynamic range.

\subsection{Comparison}

We run the AlexNet model~\cite{krizhevskyImageNetClassificationDeep2012} on CIFAR10~\cite{Krizhevsky2009LearningML} for testing both TensorFI2~\cite{liTensorFIConfigurableFault2018} and PyTorchFI~\cite{mahmoudPyTorchFIRuntimePerturbation2020}. The results for \emphasizedworkname{} are instead obtained averaging all the executions from the results, both on \glsxtrshort{cpu} and \glsxtrshort{gpu}.

\section{Evaluation}
\label{section:evaluation}

We evaluate our \emphasizedworkname{} framework based on the experiments mentioned in Section \ref{subsection:experiments}.

\subsection{Execution Time Comparison}
\label{subsection:execution_time_comparison}

\begin{figure}[t]
    \centering
    \includegraphics[width=\linewidth]{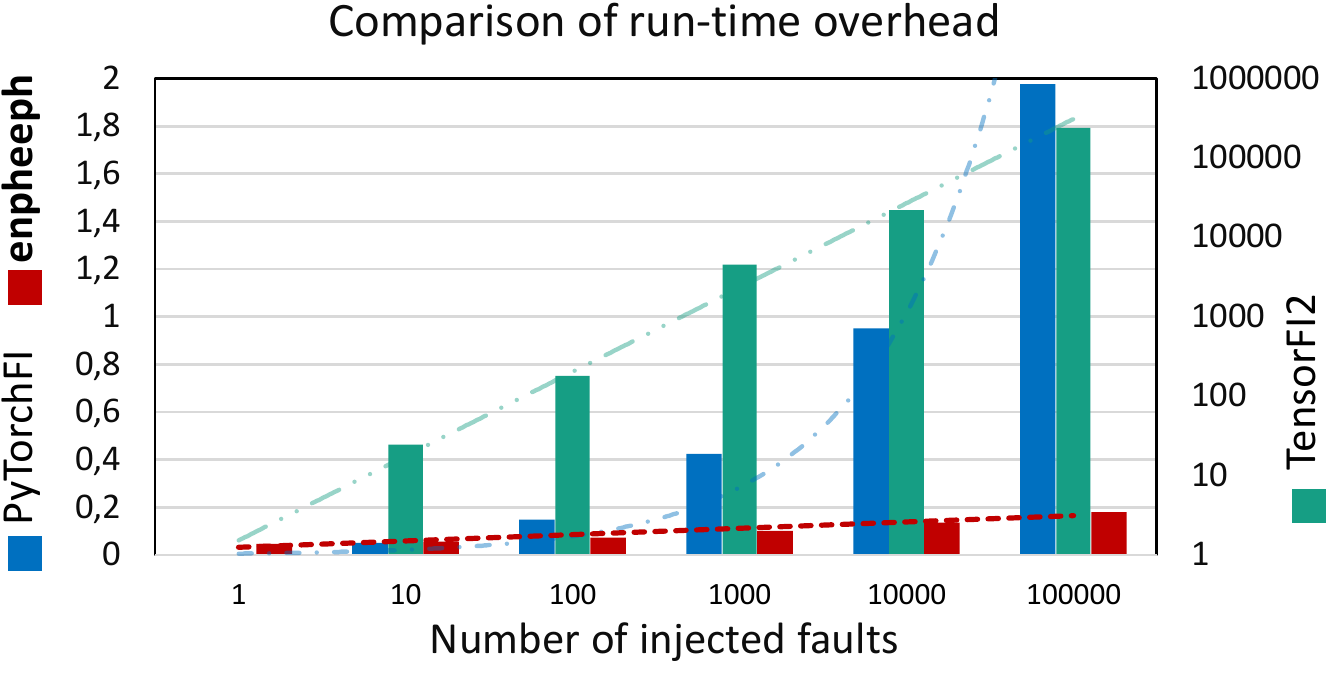}
    \caption{Comparison of run-time overhead the across state-of-the-art fault injection tools PyTorchFI and TensorFI, with our \emphasizedworkname{} framework. The overhead is measured in multiplicative units compared to the baseline, so an overhead of 0.5 indicates an execution time that is 1.5$\times$ the fault-free baseline execution time. The axis on the left is linear and refers to \emphasizedworkname{} and PyTorchFI, while the axis on the right is logarithmic and refers to TensorFI2, hence the trend is exponential even though it is represented as a linear relationship. \emphasizedworkname{} is much faster for multiple faults, and its overhead trend is linear instead of being exponential.}
    \label{figure:runtime_overhead_comparison}
\end{figure}

A comparison between \emphasis{enpheeph} and other state-of-the-art fault injection tools is shown in Fig.~\ref{figure:runtime_overhead_comparison}. We run the fault injection with up to \numprint{100000} faults, as we concurrently run multiple batches and we inject faults in multiple bits. The overhead is computed as percentage with respect to the baseline, which is the execution of the models without any injection. This scenario is realistic for faults that might happen in critical control logic, modifying many values at the same time. Our \emphasis{enpheeph} framework is much faster for multiple faults, achieving less than 20\% overhead at \numprint{100000} faults, against the 200\% of PyTorchFI and the \numprint{100000000}\% of TensorFI2. Additionally, \emphasis{enpheeph} shows a linear trend for increasing number of faults, while PyTorchFI and TensorFI2 both show exponential increases.

\subsection{\glsxtrshort*{dnn} Resiliency Analysis}

\begin{figure*}[t]
    \centering
    \includegraphics[width=\linewidth]{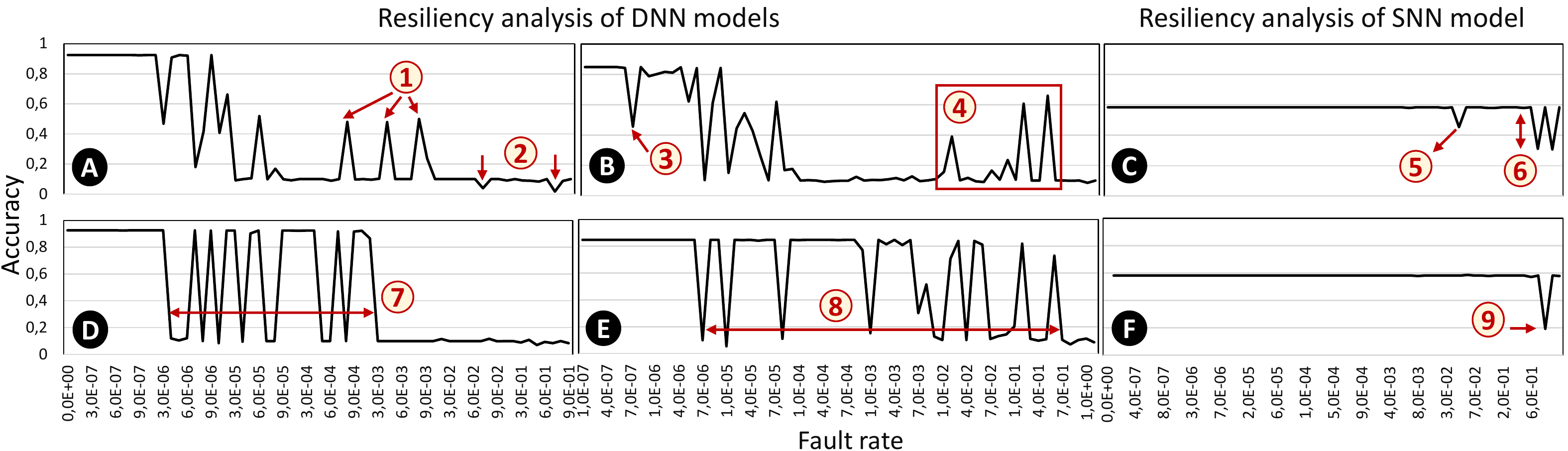}
    \caption{Comparison of testing accuracy against random faults for different models running an image classification task on CIFAR10. \GoodBlackCircled{A} shows output injection on ResNet18, \GoodBlackCircled{B} shows output injection on VGG11,
    \GoodBlackCircled{C} shows output injection on SNN,
    \GoodBlackCircled{D} shows weight injection on ResNet18,
     \GoodBlackCircled{E} shows weight injection on VGG11 and
      \GoodBlackCircled{F} shows weight injection on SNN.
     The \glsxtrshort{snn} shows higher resiliency than the \glsxtrshortpl{dnn}.}
    \label{figure:gpu_injection_image_classification_cifar10}
\end{figure*}

Analyzing the standard \glsxtrshort{dnn} models, we can see that their accuracy can be largely affected by the faults, even if random.
First, we analyze the faults injected in the layer outputs, shown in Fig.~\ref{figure:gpu_injection_image_classification_cifar10}~\GoodBlackCircled{A}\&\GoodBlackCircled{B}. As shown by~\GoodRedCircled{1} even though the accuracy drops to 10\%, which is equivalent to a random classification result, there are still some output elements that do not affect the final accuracy, as otherwise it would be impossible to obtain better than random classification with a very high fault rate, around $5 \times 10 ^{-3}$. At the same time, we can see in~\GoodRedCircled{2} that some faults force the accuracy to be below random classification, hence some elements are fundamental to the proper functioning of the model. At~\GoodRedCircled{3}, we see that VGG11 is very susceptible to faults as a fault rate as little as $7 \times 10 ^{-7}$ is to decrease the accuracy from 84\% to 45\%. However, as shown before in ResNet18, also VGG11 has some elements which are less susceptible to faults, shown in~\GoodRedCircled{4}, with peaks reaching back up to 65\% accuracy.

Regarding the layer weight injections, the results show that overall the networks are more resilient: both ResNet18 and VGG11 have a very wide range where, depending on which elements are fault-injected, the accuracy can stay close to the original model or drop to random classification levels, as shown by~\GoodRedCircled{7} and~\GoodRedCircled{8} for ResNet18 and VGG11 respectively.

When analyzing the \glsxtrshort{snn} model, shown in Fig.~\ref{figure:gpu_injection_image_classification_cifar10}~\GoodBlackCircled{C}\&\GoodBlackCircled{F}, we can notice the much higher resiliency of \glsxtrshort{snn} compared to \glsxtrshortpl{dnn}, as the first accuracy drops occurs only at $6 \times 10 ^{-2}$, as pointed by~\GoodRedCircled{5}. Additionally, when injecting layer outputs, the maximum accuracy drop is limited to just 30\%, as shown by~\GoodRedCircled{6}. When injecting weights of the \glsxtrshort{snn}, the model keeps a very high accuracy until $7 \times 10 ^{-1}$ fault rate, pointed by~\GoodRedCircled{9}.

Overall, the \glsxtrshort{snn} model proves to be more resilient than the standard \glsxtrshortpl{dnn}, which can be attributed to the time-dimension, increasing the information density and the parameter redundancy.

\subsection{Compressed networks}

We will now analyze the results of the fault injections on the layer outputs of different compressed networks, namely the \glsxtrshortpl{dnn} and the \glsxtrshort{snn} models first with 32-bit integer quantized and then with sparse indices injection. 

\begin{figure}[t]
    \centering
    \includegraphics[width=\linewidth]{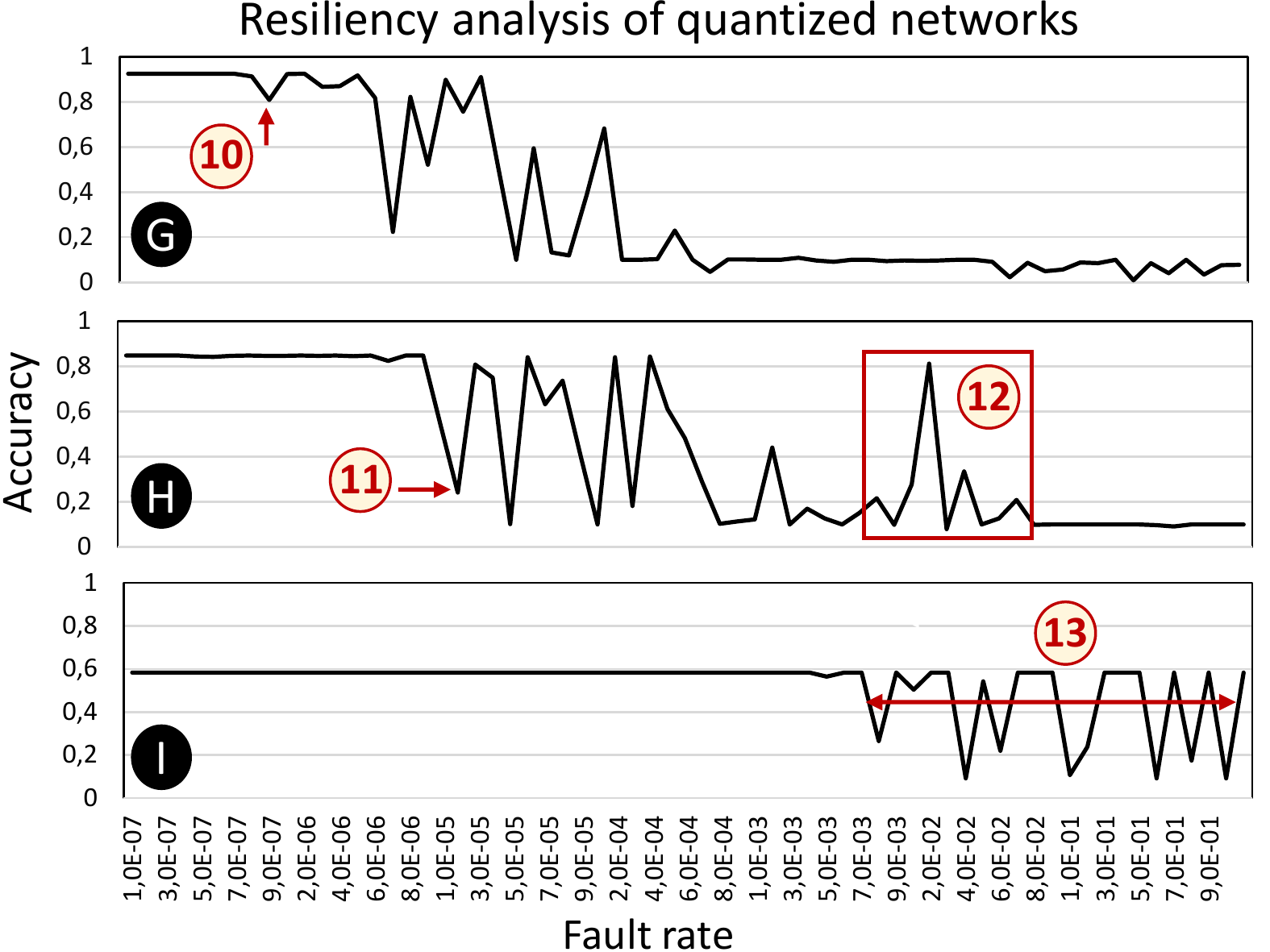}
    \caption{Testing accuracy of quantized \glsxtrshortpl{dnn} against increasing fault injection rate on the outputs. \GoodBlackCircled{G} shows ResNet18, \GoodBlackCircled{H} shows VGG11 and \GoodBlackCircled{I} shows our \glsxtrshort{snn} model. Quantization affects the resiliency with non-trivial patterns when compared to full-precision networks.}
    \label{figure:quantized}
\end{figure}

In Fig.~\ref{figure:quantized} we show the resiliency of the quantized layer outputs, for ResNet18 in \GoodBlackCircled{G}, VGG11 in \GoodBlackCircled{H} and our \glsxtrshort{snn} model in \GoodBlackCircled{I}. ResNet18 is the least resilient, with the first notable drop in accuracy around $9 \times 10 ^{-7}$, pointed by~\GoodRedCircled{10}. VGG11 has a much bigger drop around $3 \times 10 ^{-5}$, which reaches to 25\% accuracy, shown by~\GoodRedCircled{11}. However, compared to ResNet18, VGG11 has an area between $8 \times 10 ^{-3}$ and $8 \times 10 ^{-2}$, as shown by~\GoodRedCircled{12}, where the accuracy is higher than the random classification even though a higher number of fault is injected. This is related to lower-weight bits being targeted during the fault injection, leading to a lower impact on the total accuracy. For the \glsxtrshort{snn} model, accuracy is constant until $7 \times 10 ^{-3}$, where it starts alternating low accuracy and high accuracy, as pointed by~\GoodRedCircled{13}. This behaviour highlights the higher resiliency of the \glsxtrshort{snn} model, even though it shows that quantization has an effect on the resiliency of all the models. More specifically, when bits closer to the \titlecaseabbreviation{msb} are hit, the effects are increased compared to floating-point numbers, but if a bit closer to the \titlecaseabbreviation{lsb} is hit, the effects are attenuated.

\begin{figure}[t]
    \centering
    \includegraphics[width=\linewidth]{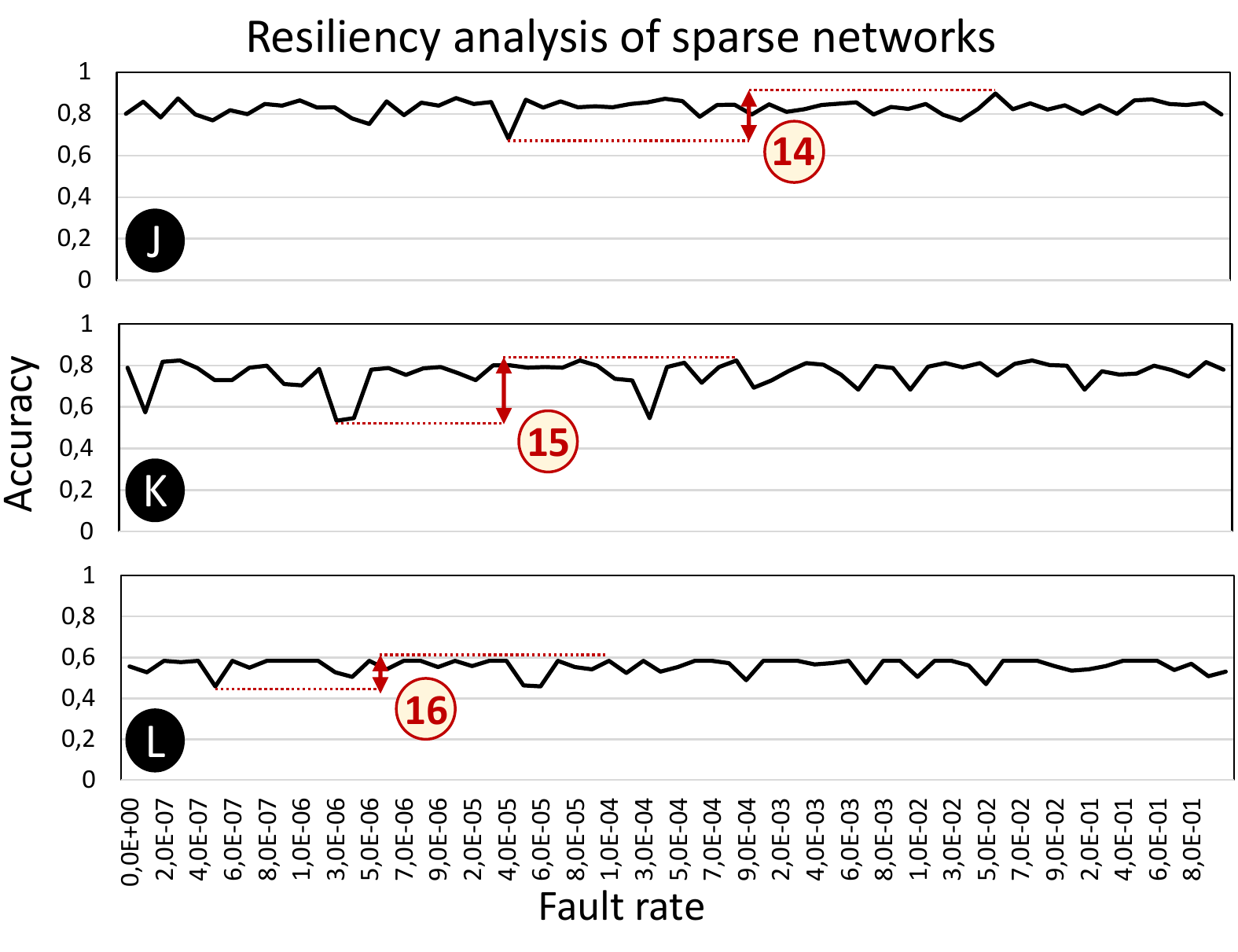}
    \caption{\GoodBlackCircled{J} shows ResNet18, \GoodBlackCircled{K} shows VGG11 and \GoodBlackCircled{L} shows our \glsxtrshort{snn} model.}
    \label{figure:sparse}
\end{figure}

Regarding sparse networks, we can see the results in Fig.~\ref{figure:sparse}~\GoodBlackCircled{J}-\GoodBlackCircled{L}. All the networks show limited effects of the injected faults, even at very high fault rates:~\GoodRedCircled{14},~\GoodRedCircled{15} and~\GoodRedCircled{16} show the range of accuracy at various fault rates for ResNet18, VGG11 and the \glsxtrshort{snn} model. The range is slightly bigger for VGG11, being roughly around 20\%. This shows that exchanging some tensor values in the output does not affect greatly the accuracy, even though further experiments are needed.\\\\

\vspace*{-10mm}
\section{Conclusion}
\label{section:conclusion}
The rising technology of neural networks for critical applications introduces the potential of many faults occurring concurrently, which is not included in state-of-the-art resiliency mitigations and analyses.
%Current state-of-the-art frameworks for resiliency analysis are designed to evaluate unique fault patterns occurring in certain \glspl{dnn}, making it impossible to analyze newer models and methodologies without incurring significant implementation delays.

To assist in analyzing \glspl{dnn} reliability, we propose \emphasis{enpheeph}, a Fault Injection Framework for Spiking and Compressed \glspl{dnn}.
In \emphasis{enpheeph}, fault injection may be executed on both \titlecaseabbreviationpl{snn} and compressed networks with little to no modification to the underlying code, a feat that other state-of-the-art tools are incapable of accomplishing.
To experiment with our \emphasis{enpheeph} framework, we examine the resilience of various \glspl{dnn} and \titlecaseabbreviationpl{snn} when compressed using various approaches.
By injecting a random and growing number of faults, we demonstrate that \glspl{dnn} have their accuracy reduced by more than 40\%, when receiving as little as $7 \times 10 ^{-7}$ random faults.
Additionally, \emphasis{enpheeph} exhibits $10 \times$ less run-time overhead than than state-of-the-art fault-injection frameworks.
We release the source code of our \emphasizedworkname~framework under an open-source license at \url{https://github.com/Alexei95/enpheeph}.

%By injecting a random and increasing number of faults, we show that \glspl{dnn} can be affected with a fault rate as low as $7 \times 10 ^{-7}$ faults per parameter, reducing the accuracy by more than 40\%. Additionally, we show the increased resiliency of \glspl{snn}, which do not show any accuracy reduction with fault rate as high as $6 \times 10 ^{-2}$. Additionally, run-time overhead when executing \emphasizedworkname is less than 20\% when executing 100'000 faults concurrently, at least $10 \times$ lower than state-of-the-art frameworks.

\section*{Acknowledgment}
\label{section:acknowledgment}

This work has been supported by the Doctoral College Resilient Embedded Systems, which is run jointly by the TU Wien’s Faculty of Informatics and the UAS Technikum Wien.

This research is partly supported by the NYUAD's Research Enhancement Fund (REF) Award on \enquote{eDLAuto: An Automated Framework for Energy-Efficient Embedded Deep Learning in Autonomous Systems}, and by the NYUAD Center for Artificial Intelligence and Robotics (CAIR), funded by Tamkeen under the NYUAD Research Institute Award CG010.

\printbibliography

\end{document}